\documentclass[letterpaper,10pt,conference]{ieeeconf}

\IEEEoverridecommandlockouts
\usepackage{graphicx}
\usepackage{amsmath}
\usepackage{amssymb}
\usepackage{booktabs}
\usepackage{url}
\usepackage{makecell}
\usepackage{xcolor}
\usepackage{booktabs}

\usepackage{multirow}
\usepackage{tabularx}
\usepackage{array}
\title{\LARGE \bf
TADreamer: Zero-Shot Language-Guided 3D Navigation for Terrestrial-Aerial Bimodal Robots via Video Imagination
}

\author{Xiangyu Li\textsuperscript{*}, Tiancheng Lai\textsuperscript{*}, Xijie Huang, Ruitian Pang, Siqi Shen, Juncheng Chen, \\
 Zaisheng Pan, Chao Xu, Fei Gao, Yanjun Cao\textsuperscript{\textdagger}%
 \thanks{\textsuperscript{*}Equal contribution. \textsuperscript{\textdagger}Corresponding author.}%
\thanks{Xiangyu Li, Tiancheng Lai, Xijie Huang, Ruitian Pang, Siqi Shen, Juncheng Chen, Chao Xu, Fei Gao and Yanjun Cao are with the State Key Laboratory of Industrial Control Technology, Zhejiang University, Hangzhou 310027, China, and also with the Huzhou Institute of Zhejiang University, Huzhou 313000, China (email: xiangyu.li@zju.edu.cn; yanjunhi@zju.edu.cn)}%
\thanks{Zaisheng Pan is with the Institute of Cyber-Systems and Control, Zhejiang University, Hangzhou 310027, China}%
}

\begin{document}

\maketitle
\thispagestyle{empty}
\pagestyle{empty}

%%%%%%%%%%%%%%%%%%%%%%%%%%%%%%%%%%%%%%%%%%%%%%%%%%%%%%%%%%%%%%%%%%%%%%%%%%%%%%%%
\begin{abstract}
Language-guided navigation for terrestrial-aerial bimodal robots requires selecting routes and locomotion modes that match scene context and task intent.
Generated videos can represent such motion sequences, but recovering metrically consistent navigation references from them is challenging because of scale ambiguity and axis-dependent geometric distortions.
We present TADreamer, a zero-shot framework that grounds video-imagined navigation in measured geometry without task-specific training or fine-tuning.
A vision-language model translates onboard observations and instructions into navigation prompts, selects valid generated videos, and provides corrective feedback when regeneration is needed.
The selected video is reconstructed into 3D waypoints annotated with terrestrial or aerial modes.
A two-stage calibration procedure uses field-of-view constraints to initialize scale estimation, then refines axis-dependent scales, rotation, and translation by registering the reconstructed point cloud to measured geometry.
The calibrated waypoints and mode labels guide a planner that incorporates measured geometry for robot execution.
Real-world experiments demonstrate navigation across seven indoor and outdoor scenarios.
With five candidates per round, usable videos are obtained within two rounds in all seven scenarios.
On the calibration observations, our method reduces mean absolute depth error by 87.7\% and mean absolute relative depth error by 86.3\% compared with NavDreamer.
\end{abstract}

%%%%%%%%%%%%%%%%%%%%%%%%%%%%%%%%%%%%%%%%%%%%%%%%%%%%%%%%%%%%%%%%%%%%%%%%%%%%%%%%
\section{Introduction}
\label{sec:introduction}

\begin{figure*}[h]
    \centering
    \vspace*{2mm}
    \includegraphics[width=0.98\textwidth]{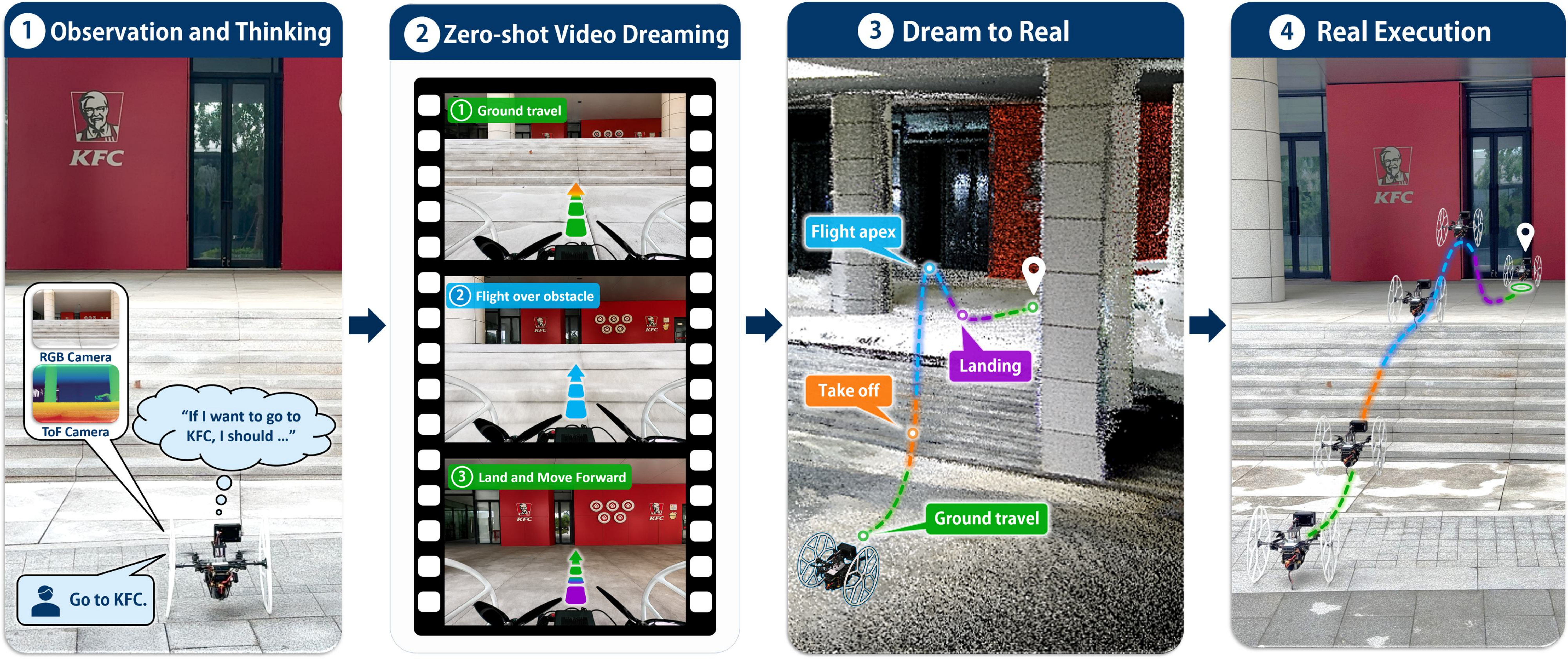}
    \caption{Language-guided terrestrial-aerial navigation with TADreamer. The instruction “Go to KFC” guides video imagination, waypoint calibration against measured geometry, and robot execution through ground travel, flight over the stairs, and landing.}
    \label{firstpicture}
    \vspace*{-4mm}
\end{figure*}

Animals in nature often adapt to complex environments by employing multiple modes of locomotion.
The roadrunner, for example, primarily relies on agile and energy-efficient terrestrial locomotion. When threatened, it can also perform short flights to avoid obstacles and evade pursuing predators.
A similar mechanism is embodied in terrestrial-aerial bimodal robots (TABRs)~\cite{lai2025trofybot,lin2024skater,li2026triphibot}. When the terrain is traversable, these robots use terrestrial locomotion to reduce energy consumption and extend operational endurance; when the ground becomes impassable, they use flight to overcome obstacles.
TABRs therefore offer distinct advantages in applications such as search and rescue, inspection, and exploration of unknown environments.
However, the key to fully exploiting the advantages of TABRs lies in deciding how terrestrial and aerial locomotion should be composed according to scene context and task instructions, rather than merely finding a single-mode collision-free path.
This requires semantic traversability reasoning, geometrically grounded decision-making, and metric consistency between imagined motion and the real environment.

Geometric representations alone are insufficient for semantic traversability reasoning.
TABR navigation requires understanding terrain traversability, obstacle properties, and language instructions.
Existing methods typically use geometric representations for collision detection and planning~\cite{fan2019autonomous,zhang2022autonomous,lai2026explore}, but these representations often lack semantic information and task intent.
For example, a flexible curtain and a rigid wall may both appear as obstacles, although the former may permit terrestrial traversal~\cite{li2026ds}.
Such ambiguity can lead to unnecessary flight or infeasible ground routes.
% Navigation decisions for TABRs should jointly consider traversable regions in the surrounding environment, the locations and types of obstacles, and the requirements specified by language instructions.
% Existing navigation methods typically rely on geometric representations, such as occupancy grids, elevation maps, or point clouds, for collision detection and path planning~\cite{fan2019autonomous,zhang2022autonomous}. Although these representations capture distance, height, and free space, they generally do not explicitly encode object properties, semantic traversability, or task intent.
% For example, both a flexible curtain and a rigid wall may appear as obstacles on a geometric map. However, the former may permit terrestrial traversal, whereas the latter must be bypassed or flown over~\cite{li2026ds}. Consequently, a purely geometric planner may either trigger unnecessary flight due to overly conservative judgments or, because of insufficient semantic understanding of the environment, attempt to traverse an impassable region on the ground.

Vision-language models (VLMs) jointly interpret visual observations and language instructions to understand task intent, reason about scenes, and generate high-level navigation decisions.
However, these decisions require geometric grounding for execution.
In aerial navigation, See, Point, Fly~\cite{hu2025see} uses a frozen VLM to identify two-dimensional waypoints in first-person images and converts them into three-dimensional motion commands through a geometric module.
FlightGPT~\cite{cai2025flightgpt} combines semantic maps with language-specified objectives to predict navigation goals.
For TABRs, DS-LABRNav~\cite{li2026ds} uses a VLM to assess obstacle traversability and guide locomotion-mode planning.
% Therefore,semantic decisions from VLMs still require geometric grounding to support metrically consistent trajectory planning.
% Second, VLM-based semantic decisions lack geometric grounding for executable navigation.
% Vision-language models (VLMs) integrate visual and linguistic information, bridging the gap between geometric perception and semantic decision-making.
% By jointly processing visual observations and natural-language instructions, VLMs can interpret task requirements, reason about environmental scenes, and generate high-level navigation decisions.
% Recent studies on aerial robot navigation have demonstrated this capability. For example, See, Point, Fly~\cite{hu2025see} uses a frozen VLM to identify two-dimensional waypoints in first-person-view images and then lifts them into three-dimensional motion commands through a geometric module. FlightGPT~\cite{cai2025flightgpt} employs a VLM to integrate semantic maps with language-specified objectives and predict navigation goals.
% For TABRs, DS-LABRNav~\cite{li2026ds} further uses a VLM to assess obstacle traversability and guide locomotion-mode planning accordingly.
% However, semantic decisions from VLMs still require geometric grounding to support metrically consistent trajectory planning.

Generative video models offer a bridge between semantic decision-making and spatial motion.~\cite{du2023learning,du2024video}
Conditioned on visual observations and task instructions, these models can synthesize imagined navigation sequences, from which camera poses can be recovered to construct candidate trajectories.
NavDreamer~\cite{huang2026navdreamer} demonstrates the potential of this approach for zero-shot 3D navigation. However, visually plausible videos do not necessarily produce geometrically consistent or executable trajectories.
Monocular reconstruction remains scale-ambiguous without metric priors, while video generation may introduce scene hallucinations and non-uniform geometric distortions.
Here, axis-dependent geometric distortion refers to anisotropic scaling in the reconstructed geometry, where distances along different spatial axes are distorted by different factors.
Consequently, a single global scale factor cannot simultaneously align all axes, leading to systematic errors in the recovered waypoints.
These limitations motivate calibrating imagined waypoints against measured geometry to provide metrically aligned references for downstream trajectory planning.
% Generative video models provide a new bridge between semantic decision-making and spatial motion.
% Given the current visual observation and task instruction, a video model can imagine a video consistent with the specified objective, from which camera poses can then be estimated and used to construct a motion trajectory.
% NavDreamer~\cite{huang2026navdreamer} has demonstrated the potential of video generation models for zero-shot 3D navigation.
% However, visually plausible imagination does not necessarily yield an executable motion trajectory. video-imagined motion lacks metric consistency with the real environment.
% Without metric priors, monocular reconstruction is scale-ambiguous, while video generation may introduce scene hallucinations and non-uniform distortions.
% Moreover, a single global scale-calibration factor cannot adequately compensate for anisotropic errors along different axes.
% Therefore, calibrating error-prone waypoints recovered from imagined scenes into executable waypoints in the real environment is critical to leveraging generative video.

To address the aforementioned challenges, we propose a scene-aware video navigation framework for TABRs.
The framework employs a VLM to interpret first-person robot observations and natural-language instructions and generate prompts that guide imagined-video generation, without requiring manually designed decomposed prompts.
After the VLM selects a generated video and identifies its locomotion modes, the Depth Anything 3 (DA3)~\cite{lin2025depth} decoder recovers the camera trajectory and imagined point cloud.
The imagined point cloud is then registered with the measured point cloud to estimate the transformation parameters and correct the metric-scale deviations of the recovered waypoints.
Finally, the aligned waypoints are passed to a low-level planner, which generates a continuous trajectory satisfying collision-avoidance and motion constraints for execution by the robot.
Fig.~\ref{firstpicture} illustrates a representative task in which TADreamer follows a language instruction through ground travel, flight over stairs, and landing and rolling to the destination.
The main contributions of this work are as follows:
\begin{itemize}
    % \item We develop a 3D navigation framework for TABRs that integrates a VLM with a video generation model, combining environmental perception and instruction understanding to enable an "imagine-then-act" navigation paradigm.
    % \item We propose a calibration method using measured point clouds as metric references, converting waypoints decoded from imagined video into executable waypoints consistent with the scale of the real environment.
    % \item We conduct real-world and simulation experiments to evaluate the navigation performance of semantic decision-making, video generation, and waypoint calibration.
    % \item We develop a language-guided navigation framework for TABRs that represents route progression, locomotion-mode transitions, and stopping conditions through generated videos and converts them into planning references.
    % \item We design a two-stage calibration procedure that uses measured point clouds to initialize and refine axis-dependent scales, aligning video-reconstructed geometry and waypoints with the measured environment.
    % \item We evaluate the framework in seven real-world scenarios, examining video usability, waypoint-mode annotations, and reconstructed depth accuracy.
    \item We introduce TADreamer, a zero-shot language-guided navigation framework for TABRs that uses VLM-guided video imagination to compose terrestrial and aerial locomotion according to scene context and task instructions.
    \item We design a two-stage calibration procedure that uses measured point clouds to initialize and refine axis-dependent scales, aligning video-reconstructed geometry and waypoints with the measured environment.
    \item We validate TADreamer in seven real-world scenarios, demonstrating usable multimodal navigation videos and substantially improved reconstruction accuracy over existing baselines. 
\end{itemize}

%%%%%%%%%%%%%%%%%%%%%%%%%%%%%%%%%%%%%%%%%%%%%%%%%%%%%%%%%%%%%%%%%%%%%%%%%%%%%%%%
\section{Related Work}
\label{sec:related_work}
\subsection{Autonomous Navigation for TABRs}
TABRs trade ground endurance for aerial reachability, so navigation must choose both a route and a locomotion mode.
Fan et al.~\cite{fan2019autonomous} unified rolling and flying control with a differential-flatness planner for unknown environments.
Zhang et al.~\cite{zhang2022autonomous} coupled hierarchical kinodynamic search with B-spline optimization to account for ground curvature and flight energy.
Their later model-based framework integrated unified dynamics, trajectory optimization, and nonlinear model predictive control for both modes~\cite{zhang2023model}.
Li et al.~\cite{li2025two} reduced online computation by predicting global land--air keypoints from depth and goal inputs before map-based local refinement.
For exploration, Gao et al.~\cite{gao2025autonomous} jointly scheduled bimodal viewpoints and modes under time and energy budgets.
These approaches provide geometric planning and control under motion and resource constraints.
However, these geometry-planning-based methods only demonstrate the motion capabilities of TABRs, falling short of effectively leveraging natural language instructions and environmental semantics in real-world settings.

\subsection{VLM-Based Semantic and Affordance-Aware Navigation}
VLM-based navigation augments geometric perception with goal semantics and reasoning about scene affordances~\cite{yokoyama2024vlfm}.
LM-Nav~\cite{shah2023lm} decomposes free-form instructions into landmark sequences, grounds them with CLIP, and connects them using a learned navigation policy.
VLMaps~\cite{huang2023visual} fuses pretrained vision--language features into a 3D map, enabling open-vocabulary object and spatial-relation queries.
For UAVs, See, Point, Fly~\cite{hu2025see} uses a frozen VLM to predict image-plane waypoints and travel distances, then lifts them to 3D displacements with geometry.
FlightGPT~\cite{cai2025flightgpt} reasons over a semantic map and language goal with look-ahead planning.
For a single TABR, DS-LABRNav~\cite{li2026ds} invokes a VLM before takeoff to assess obstacle traversability and update the occupancy map.
These systems primarily translate semantic information into landmarks, navigation goals, control commands, or map updates.
A remaining challenge is to represent route progression, locomotion-mode transitions, and stopping conditions jointly within a temporally coherent predictive representation.

\subsection{Generative Video Models for Robot Planning}
Generative video models provide a visual planning space in which predicted observations encode long-horizon behavior. UniPi generates text-conditioned plans and converts them to controls with an inverse-dynamics model~\cite{du2023learning}.
Video Language Planning combines a video dynamics model, VLM-based policy/value functions, and tree search, then executes selected futures through goal-conditioned policies~\cite{du2024video}.
Recent navigation systems bring this idea closer to deployment. NavDreamer~\cite{huang2026navdreamer} ranks generated videos with a VLM, reconstructs 3D waypoints, and estimates a single global scale from monocular metric-depth priors; ImagiNav~\cite{chen2026imaginav} couples language subgoals, visual prediction, and inverse dynamics for embodied navigation.
DreamToNav~\cite{serpiva2026dreamtonav} refines language prompts with a VLM and recovers executable robot trajectories from third-person generated videos using robot detection, visual odometry, and pose estimation.
Action Agent~\cite{sam2026action} adds generator selection, iterative video validation, prompt revision, and flow-conditioned action prediction.
These approaches connect visual imagination to action, but a generated trajectory still requires a reliable metric interface for execution on a hybrid platform.
%%%%%%%%%%%%%%%%%%%%%%%%%%%%%%%%%%%%%%%%%%%%%%%%%%%%%%%%%%%%%%%%%%%%%%%%%%%%%%%%
\begin{figure*}[h]
    \centering
    \vspace*{2mm}
    \includegraphics[width=0.96\textwidth]{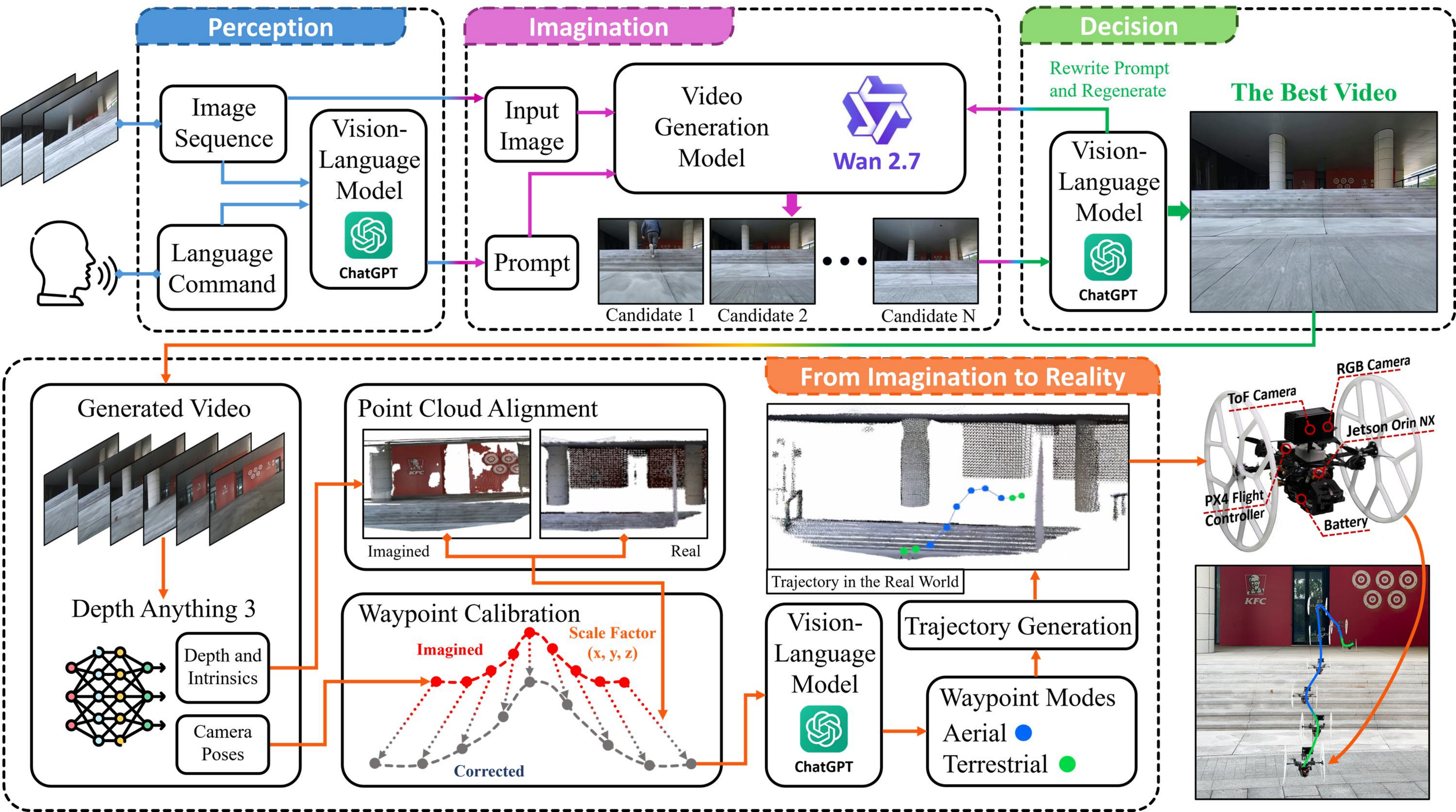}
    \caption{Overview of TADreamer. A VLM generates prompts, selects videos with corrective feedback, and assigns locomotion modes. DA3 reconstructs imagined geometry, while measured point clouds calibrate the recovered waypoints. A mode-aware planner generates executable trajectories. Green and blue denote terrestrial and aerial waypoints.}
    \label{pipeline}
\end{figure*}
\begin{figure}[h]
    \centering
    \vspace*{2mm}
    \includegraphics[width=0.96\columnwidth]{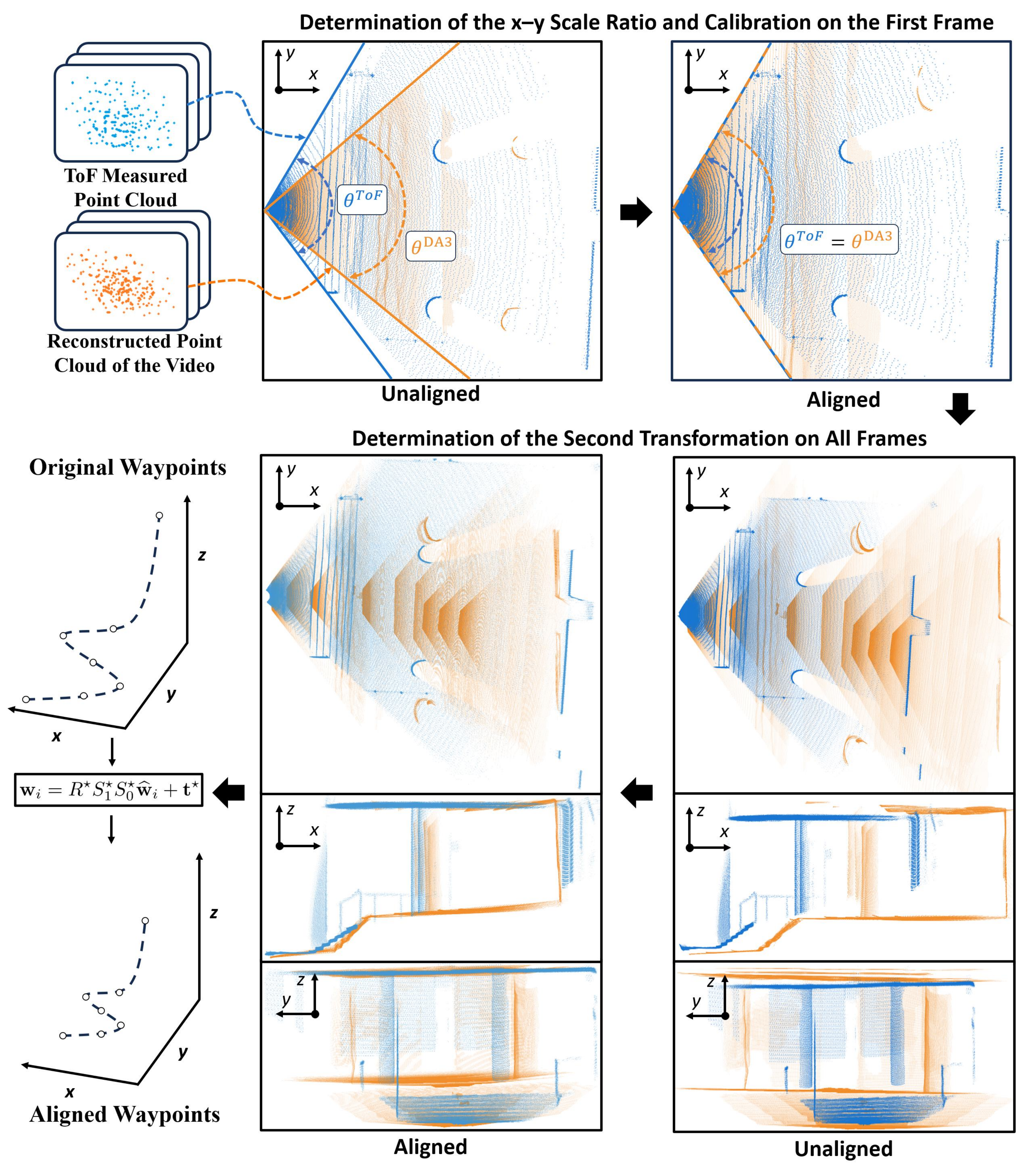}
    \caption{Two-stage calibration using FoV-based scale initialization and full-cloud anisotropic registration. The estimated transformation corrects recovered waypoints. Blue and orange denote measured and imagined point clouds, respectively.}
    \label{matching}
\end{figure}
\section{Method}
\label{sec:method}
Our framework comprises VLM-guided video generation and selection, point cloud reconstruction, geometric calibration, and mode-aware trajectory planning, as illustrated in Fig.~\ref{pipeline}. 
In this context, zero-shot denotes the direct application of pretrained VLM, video-generation, and DA3 models without task-specific training or fine-tuning.
\subsection{VLM-Based Perception and Navigation Prompt Generation}
Brief language instructions typically specify only the task intent, whereas video generation requires an explicit description of the motion sequence appropriate for the scene.
We use a VLM to process an egocentric RGB observation clip $\mathcal{I}^{\mathrm{obs}}$, the user instruction $\ell$, and a predefined semantic prompt $P_{\mathrm{sem}}$ to generate a scene-specific navigation prompt:
\begin{equation}
P_{\mathrm{nav}}
= F_{\mathrm{VLM}}\left(\mathcal{I}^{\mathrm{obs}}, \ell; P_{\mathrm{sem}}\right)
= \left(d_{\mathrm{route}}, d_{\mathrm{mode}}, d_{\mathrm{stop}}\right).
\end{equation}
$d_{\mathrm{route}}$ specifies the motion direction, target, and sequence of motion stages; $d_{\mathrm{mode}}$ specifies the choice between ground locomotion and flight and the route segments to which each applies; and $d_{\mathrm{stop}}$ describes the state at task completion.
The three components are combined into a length-constrained prompt for video generation.

\subsection{Navigation Video Generation with VLM Feedback}

Previous studies have reported that closed-source models outperform open-source models in terms of physical plausibility and instruction adherence~\cite{huang2026navdreamer}.
We therefore adopt the Wan2.7 image-to-video model, using the first frame \(I_0\) of the current observation clip as the visual condition and combining the navigation prompt \(P_{\mathrm{nav}}\) with a predefined generation prompt \(P_{\mathrm{gen}}\) as the generation input.
\(P_{\mathrm{gen}}\) specifies visual constraints that apply to all tasks, including maintaining a first-person perspective, preserving scene structure, and avoiding camera cuts and physically implausible zooming.

Let \(r\in\{1,\ldots,R_{\max}\}\) index the generation-and-evaluation rounds.
Each round generates \(n\) candidate videos using different random seeds and evaluates them with the VLM.
The initial corrective prompt is empty, \(P_{\mathrm{fix}}^{(1)}=\varnothing\).
A new round is initiated only when all candidates are rejected and the corrective prompt is updated.
The \(j\)-th candidate in round \(r\) is generated as
\begin{equation}
V_j^{(r)} =
G_{\mathrm{wan}}\left(
I_0,\,
P_{\mathrm{nav}} \oplus P_{\mathrm{gen}} \oplus P_{\mathrm{fix}}^{(r)},
\, \zeta_j^{(r)},\, \Theta_G
\right),
\end{equation}
\(\oplus\) denotes prompt combination, \(\zeta_j^{(r)}\) is the random seed, \(\Theta_G\) denotes the generation parameters, and \(P_{\mathrm{fix}}^{(r)}\) is the corrective prompt for the current round, which is initially empty.
Since video generation is stochastic, a single sample may yield a video of poor quality or one that fails to meet the requirements.
We therefore generate candidate videos using multiple random seeds to reduce the impact of random variation.

Given the candidate videos, the VLM evaluates their validity and identifies requirement violations using the navigation prompt and a predefined evaluation prompt \(P_{\mathrm{eval}}\):
\begin{equation}
\left\{
\left(b_j^{(r)}, \mathcal{E}_j^{(r)}\right)
\right\}_{j=1}^{n}
=
F_{\mathrm{VLM}}\left(
\left\{V_j^{(r)}\right\}_{j=1}^{n},
P_{\mathrm{nav}}, P_{\mathrm{eval}}
\right).
\end{equation}
The validity indicator \(b_j^{(r)}\in\{0,1\}\) specifies whether the candidate follows the prescribed route, locomotion modes, and stopping condition while maintaining a continuous first-person view and consistent scene structure.
The violation set \(\mathcal{E}_j^{(r)}\) records specific failures, such as unexpected object motion, newly introduced actors, or zoom-only apparent motion.

If valid candidates are available, the VLM compares their camera-motion smoothness, consistency of the prescribed motion sequence, and clarity of progress toward the target, and selects a preferred video \(V_{best}\).

If all candidates fail the evaluation, the system updates the corrective prompt based on the failure reasons:
\begin{equation}
P_{\mathrm{fix}}^{(r+1)}
=
F_{\mathrm{update}}\left(
P_{\mathrm{fix}}^{(r)},
\bigcup_{j=1}^{n} \mathcal{E}_j^{(r)}
\right).
\end{equation}
This update transforms the observed problems into specific generation constraints, while retaining the original task objective and proceeding to the next generation round.
To bound the computational cost, the maximum number of generation rounds is set to \(R_{\max}\).

After selecting the video, the landing and takeoff events of the video are obtained through VLM analysis.
\begin{equation}
\mathcal{M}^{\star}
=
F_{\mathrm{mode}}\left(
    V_{best}
    \right)
=
\left\{
\left(\tau_k^{\mathrm{fly}}, \tau_k^{\mathrm{land}}\right)
\right\}_{k=1}^{N_s}.
\end{equation}
\(\tau_k^{\mathrm{fly}}\) and \(\tau_k^{\mathrm{land}}\) denote the time of lift-off and the time at which landing is completed on the video timeline, respectively.
If landing does not occur within the video, the aerial interval extends to the final frame.
These timestamps are used to delineate locomotion modes and do not directly determine the timing of the robot's actual takeoffs and landings.

\subsection{Two-Stage Scale Calibration based on Measured Point Cloud}
The selected imagined video is sampled at fixed temporal intervals to obtain an image sequence \(\mathcal{I}=\{I_i\}_{i=0}^{K-1}\). The sequence is then fed into DA3 to recover the camera pose and scene point cloud of each frame.
To construct a unified geometric representation, we use the first frame's camera coordinate system as the reference frame, transforming the point clouds recovered from all frames into it before merging:
\begin{equation}
\widehat{\mathcal{P}}^{0}
=
\bigcup_{i=0}^{K-1}
T_{0\leftarrow i}\widehat{\mathcal{P}}_i.
\end{equation}
Here, \(\widehat{\mathcal{P}}_i\) denotes the point cloud recovered from the \(i\)-th frame in its camera coordinate system, \(T_{0\leftarrow i}\) denotes the transformation from the camera coordinate system of the \(i\)-th frame to that of the first frame, computed from the recovered camera poses, and \(\widehat{\mathcal{P}}^{0}\) denotes the merged imagined point cloud.

After coordinate alignment, the camera center of the first frame is located at the origin, while the camera centers of the remaining frames form a sequence of candidate waypoints relative to the starting position.
Each waypoint is assigned a terrestrial or aerial mode label according to the timestamp of its corresponding sampled frame in the generated video.

To correct scale discrepancies in the reconstructed point clouds and waypoints, we perform the two-stage registration illustrated in Fig.~\ref{matching}, using the measured metric point cloud as the reference.
The first stage estimates an initial scale transformation for the subsequent iterative closest point (ICP) optimization~\cite{besl1992method}.

In the first stage, the horizontal angular coverage of the measured point cloud and the horizontal field of view inferred from the first imagined frame are used to determine the planar axis ratio \(\kappa=s_{0y}/s_{0x}\), where the \(x\)- and \(y\)-axes span the horizontal plane and the \(z\)-axis represents the vertical direction.
Scaling ICP extends point-set registration to account for axis-dependent scales~\cite{du2010scaling}.
The first frames of the imagined and measured point clouds share the same position and are free of rotation.
With the FoV-derived ratio fixed, we use a constrained scale-only ICP procedure to estimate one horizontal scale parameter and one independent vertical scale parameter.
The procedure alternates between updating nearest-neighbor correspondences and optimizing the scale parameters with the correspondences held fixed.
We minimize the trimmed root-mean-square correspondence distance without applying rotation or translation, yielding
\begin{equation}
\begin{aligned}
S_0^\star
&=
\operatorname{diag}
\left(s_{0x}^\star,s_{0y}^\star,s_{0z}^\star\right), \
s_{0y}^\star
&=
\kappa s_{0x}^\star.
\end{aligned}
\end{equation}
The initial scale transformation \(S_0^\star\) is applied to the complete imagined point cloud to reduce the scale mismatch and provide an initialization for the second stage.

Starting from this initialization, the second stage removes the horizontal axis-ratio constraint and jointly refines all three axis scales, rotation, and translation within the ICP framework.
At each iteration, nearest-neighbor correspondences are recomputed under the current transformation and held fixed while updating the transformation parameters.
We define the incremental scale matrix for the second stage as
\begin{equation}
S_1
=
\operatorname{diag}
\left(s_{1x},s_{1y},s_{1z}\right).
\end{equation}
For a fixed set of correspondences, the registration objective is formulated as
\begin{equation}
\left(R^\star,\mathbf{t}^\star,S_1^\star\right)
=
\underset{R,\mathbf{t},S_1}{\operatorname{arg\,min}}
\sum_{j=1}^{N}
\left\|
R S_1\mathbf{p}_j^{(0)}
+\mathbf{t}
-\mathbf{q}_j
\right\|_2^2.
\end{equation}
Here, \(\mathbf{p}_j^{(0)}=S_0^\star\widehat{\mathbf{p}}_j\) denotes an imagined point after the initial scale transformation, \(\mathbf{q}_j\) denotes its corresponding measured point, and \(N\) is the number of valid point correspondences. The rotation matrix satisfies \(R\in SO(3)\), the translation vector is \(\mathbf{t}\in\mathbb{R}^3\), and the incremental scale factors satisfy \(s_{1x},s_{1y},s_{1z}>0\), with no equality constraint between \(s_{1x}\) and \(s_{1y}\). The registration is completed by iteratively updating the point correspondences and transformation parameters.

Finally, the composite scale transformation is defined as \(S^\star=S_1^\star S_0^\star\), and the calibrated candidate waypoints are obtained as
\begin{equation}
\begin{aligned}
\mathbf{w}_i
&=
R^\star S_1^\star S_0^\star
\widehat{\mathbf{w}}_i
+\mathbf{t}^\star \\
&=
R^\star S^\star
\widehat{\mathbf{w}}_i
+\mathbf{t}^\star.
\end{aligned}
\end{equation}
\(\widehat{\mathbf{w}}_i\) denotes the original recovered waypoint, while \(\mathbf{w}_i\) denotes the calibrated waypoint in the coordinate system of the measured point cloud. Its terrestrial or aerial mode label remains unchanged.
\subsection{Mode-Aware Trajectory Generation}
Even after calibration against the measured point cloud, the waypoints recovered from the generated video may still contain local nonlinear geometric errors, particularly in regions that are occluded from the initial viewpoint, such as areas behind obstacles.
Tracking these waypoints directly may still cause collision risks and non-smooth motion.
To address this issue, we employ the TABR trajectory generator proposed by Zhang et al.~\cite{zhang2022autonomous}, which uses the aligned waypoints and their associated locomotion modes as references and incorporates the measured environmental map to generate a smooth trajectory that satisfies obstacle-avoidance and motion constraints.

In addition, in the terrestrial mode, the robot is subject to a nonholonomic no-side-slip constraint, requiring its yaw angle to align with the direction of its planar velocity during locomotion.
In the aerial mode, the yaw angle can be controlled independently of the translational motion.
Therefore, we only use the yaw angle corresponding to the recovered waypoints as a reference in the aerial mode, in order to meet the orientation requirements of specific tasks.
Finally, the planned trajectory is passed to the low-level controller for tracking and execution.
%%%%%%%%%%%%%%%%%%%%%%%%%%%%%%%%%%%%%%%%%%%%%%%%%%%%%%%%%%%%%%%%%%%%%%%%%%%%%%%%
\section{Experiments}
\label{sec:experiments}
\subsection{Real-World Navigation Experiments}

We evaluated the framework in seven indoor and outdoor scenarios using a TABR similar to that of Li et al.~\cite{li2026triphibot}. An onboard Odin1 module provides RGB images, time-of-flight (ToF) point clouds, and robot pose estimates.

The VLM and video model are accessed through APIs for prompt generation, video synthesis, and candidate selection. DA3 runs on an NVIDIA RTX 4080 at \(504 \times 406\) resolution to reconstruct camera poses and point clouds. Calibrated waypoints and locomotion-mode labels are transmitted to the onboard Jetson Orin NX, where the planner incorporates live Odin1 point clouds to generate trajectories satisfying obstacle-avoidance and motion constraints. Control commands are then sent to the PX4 flight controller.

\begin{figure}[htbp]
    \centering
        \vspace*{2mm}
    \includegraphics[width=0.99\columnwidth]{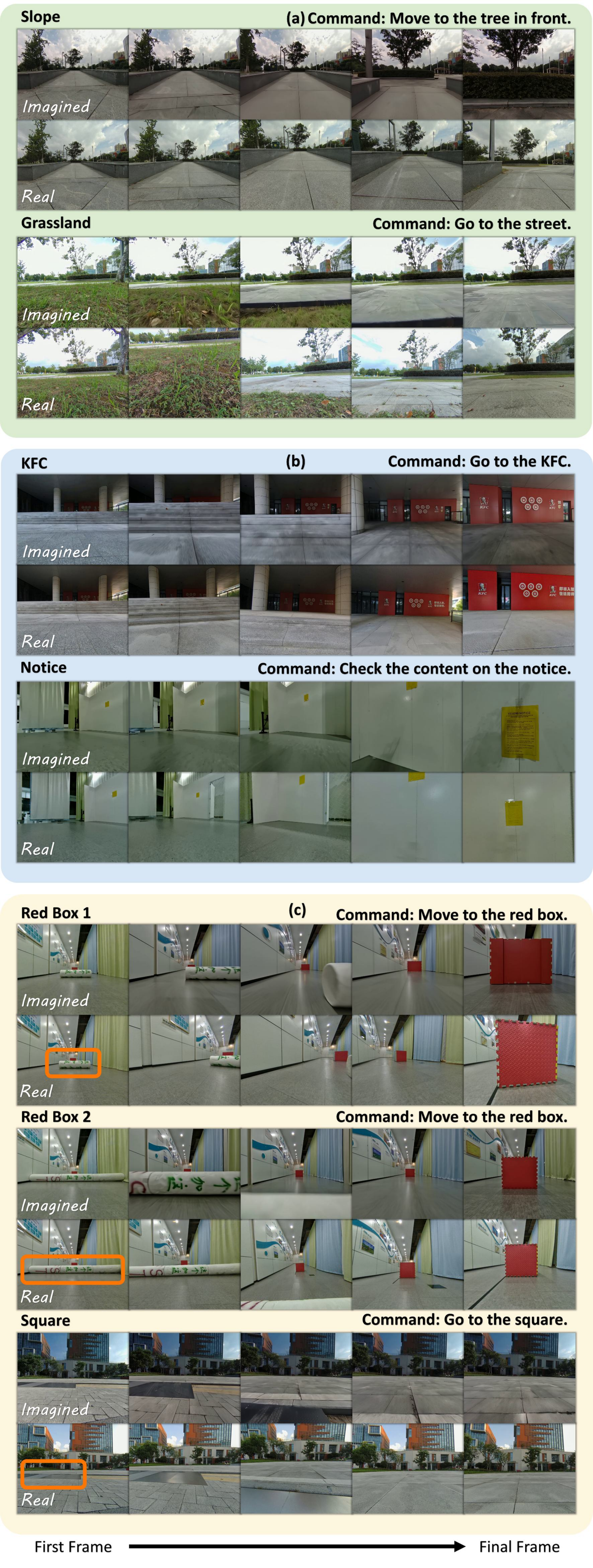}
    \caption{Imagined navigation and real-world execution across seven scenarios. Panels illustrate (a) ground traversability, (b) target-dependent terminal modes, and (c) obstacle-aware route and mode selection.}
    \label{experiment1}
\end{figure}

Fig.~\ref{experiment1} illustrates real-world execution in \textit{Slope}, \textit{Grassland}, \textit{KFC}, \textit{Notice}, \textit{Red Box 1}, \textit{Red Box 2}, and \textit{Square}. In the demonstrated trials, the robot follows the main imagined route and reaches the instructed goal.

\subsection{VLM-Based Navigation Prompt Generation}
\begin{figure}[htbp]
    \vspace*{2mm}
    \centering
    \includegraphics[width=0.99\columnwidth]{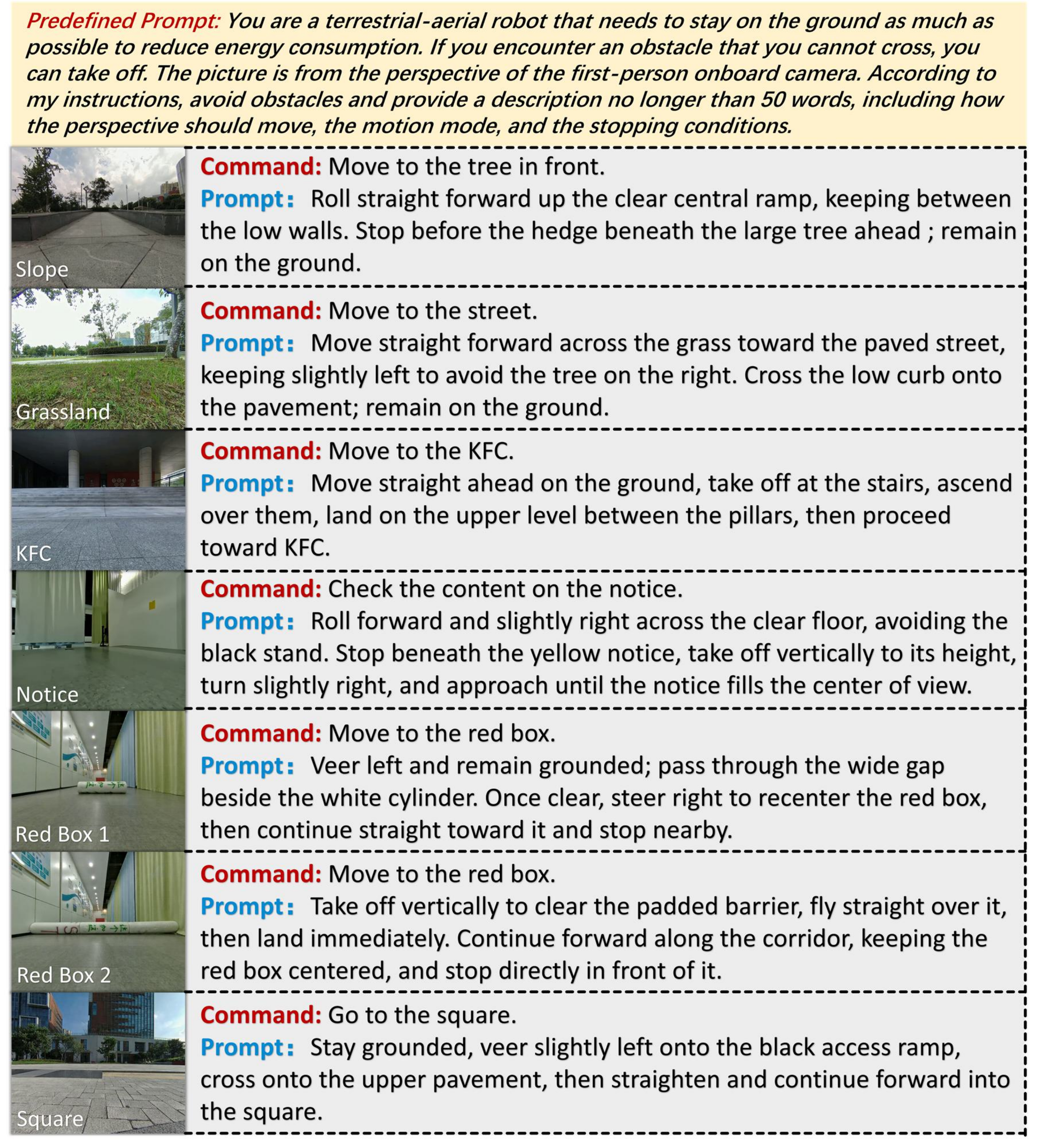}
    \caption{VLM-generated navigation prompts across seven scenarios. Each row pairs an observation and user command with a generated prompt specifying routes, locomotion modes, and stopping conditions. The shared predefined prompt encourages ground locomotion.}
    \label{prompt}
\end{figure}

We evaluated ChatGPT Sol-5.6 in seven challenging scenarios to assess whether it can combine visual observations, user instructions, and predefined prompts into navigation prompts suitable for video generation.
As shown in Fig.~\ref{prompt}, these scenarios cover terrain understanding, target recognition, path selection, and locomotion-mode decisions.

In \textit{Slope} and \textit{Grassland}, geometric planners relying on a flat-ground assumption may classify slopes or grass as non-traversable and trigger flight.
The VLM instead recognizes the terrain and explicitly specifies continued ground locomotion, using environmental semantics to avoid unnecessary takeoffs.

\textit{KFC} and \textit{Notice} require different terminal locomotion modes.
Guided by user instructions, the VLM locates the target, describes forward and turning motions, and specifies whether to remain airborne or on the ground upon arrival.
These scenarios assess its joint understanding of target semantics, motion sequences, and terminal states.

In \textit{Red Box 1}, \textit{Red Box 2}, and \textit{Square}, the VLM selects routes and locomotion modes based on obstacle distribution and the predefined energy-saving requirement.
It specifies ground detours in \textit{Red Box 1} and \textit{Square}, and takeoff only in \textit{Red Box 2}, where flight is required to cross the obstacle. This reflects a preference for ground locomotion, with flight used only when necessary.

These results indicate that, in the tested scenarios, the VLM translates visual information and language-based task requirements into navigation prompts specifying routes, locomotion modes, and terminal conditions, providing semantic guidance for subsequent video generation.
\subsection{Video Generation and Mode Annotation}
For each scene, we generated five 5-second videos per round using different random seeds and sampled frames every 0.5~s. The VLM selected a valid candidate or provided corrective feedback for regeneration.
VLM-predicted takeoff and landing times determined the locomotion-mode label of each sampled frame.
To evaluate video usability and the accuracy of the locomotion-mode annotations, we invited experienced robot operators to manually assess the videos generated in each round and verify the mode annotations of the selected videos.
In Table~\ref{tab:scene_results}, \textit{Usable videos} denotes the number of videos in each round that satisfy the basic requirements.
\textit{Failure reason} records the primary failure reason when no usable video is obtained in the first round.
\textit{Mode Success} indicates whether all locomotion-mode annotations in the selected video are correct.
A case is counted as successful only when the predicted mode of every sampled waypoint matches the reference annotation provided by the robot operators.

The results show that five of the seven scenarios yielded at least one usable video after the first generation round.
In the \textit{KFC} and \textit{Red Box 2} scenarios, none of the first-round candidates satisfied the requirements. After adding ``No people or moving objects.'' and ``All objects are stationary and do not move'' as \(P_{\mathrm{fix}}\) for the \textit{KFC} and \textit{Red Box 2} scenarios, respectively, usable videos were obtained for both scenarios in the second round.
Across the final videos selected for all seven scenarios, the locomotion modes assigned by the VLM to all sampled waypoints agreed with the reference annotations provided by the robot operators.

In addition, we found that in the experiments, the VLM was able to identify cases in the \textit{KFC} and \textit{Red Box 2} scenes where none of the candidate videos met the basic requirements, and to provide targeted correction prompts. When qualified videos were available, the VLM could also complete the selection and choose one video, but its top choice did not always agree with the the robot operators' selection.

These results indicate that the generation pipeline combining VLM-based candidate selection with prompt correction can produce usable navigation videos and provide valid locomotion-mode labels for the extracted waypoints.
\begin{table}[htbp]
    \centering
    \caption{Success and regeneration results across scenes.}
    \label{tab:scene_results}
    \begin{tabular}{lccc}
        \hline
        Scene & \makecell{Usable\\Videos}
              & \makecell{Failure\\Reason}
              & \makecell{Mode\\Success} \\
        \hline
        \textit{Slope}     & 4/5                      & None                       & Yes \\
        \textit{Grassland} & 2/5                      & None                       & Yes \\
        \textit{KFC}       & 0/5 (Round 1) & External object entering & --- \\
                         & 5/5 (Round 2) & None & Yes \\
        \textit{Notice}    & 5/5                      & None                       & Yes \\
        \textit{Red Box 1} & 2/5                      & None                       & Yes \\
        \textit{Red Box 2} & 0/5 (Round 1) & Obstacle movement        & --- \\
         & 1/5 (Round 2) & None        & Yes \\
        \textit{Square}    & 2/5                      & None                       & Yes \\
        \hline
    \end{tabular}
\end{table}

\subsection{Evaluation of Scale Calibration}
Metric-scale errors in recovered waypoints can compromise downstream trajectory planning and execution.
Although monocular geometry models can provide scale priors~\cite{hu2024metric3d}, generated videos may exhibit unaligned and time-varying camera intrinsics as well as nonlinear geometric distortions.
Consequently, the directly recovered geometry may still deviate from the true metric scale.
To evaluate this issue, we compare the scales directly recovered by DA3, the scale calibration results of NavDreamer, which uses $\pi^3$~\cite{wang2026pi} reconstruction followed by MoGe2~\cite{wang2026moge} metric depth estimation to unify scale, and the calibration results of our method across seven scenarios.

From the first-person perspective, we compute the mean absolute depth error (MADE) and the mean absolute relative depth error (MARDE) over the common valid-pixel region between the predicted depth and the ground-truth depth acquired by the onboard ToF camera:

\begin{equation}
\mathrm{MADE}
=
\frac{1}{|\Omega|}
\sum_{\mathbf{u}\in\Omega}
\left|
D_{\mathrm{pred}}(\mathbf{u})
-
D_{\mathrm{gt}}(\mathbf{u})
\right|,
\end{equation}

\begin{equation}
\mathrm{MARDE}
=
\frac{1}{|\Omega|}
\sum_{\mathbf{u}\in\Omega}
\frac{
\left|
D_{\mathrm{pred}}(\mathbf{u})
-
D_{\mathrm{gt}}(\mathbf{u})
\right|
}{
D_{\mathrm{gt}}(\mathbf{u})
}
\times 100\%.
\end{equation}
\(\Omega\) denotes the set of valid pixels, while \(D_{\mathrm{pred}}\) and \(D_{\mathrm{gt}}\) denote the predicted and ground-truth depths, respectively.
Depth errors are evaluated on the observations used for calibration and therefore quantify alignment accuracy on those observations.

As shown in Table~\ref{tab:scale_calibration}, our method outperforms DA3 and NavDreamer in all evaluated scenarios.
Compared with DA3 and NavDreamer, our method reduces the mean absolute depth error by \(80.0\%\) and \(87.7\%\), respectively, and the mean absolute relative depth error by approximately \(75.5\%\) and \(86.3\%\), respectively.
Comparisons across scenarios further show that the geometric reconstruction errors in indoor, short-range scenarios, such as \textit{Notice} and \textit{Red Box 1}, are lower than those in outdoor, long-range scenarios, such as \textit{Grassland} and \textit{Square}.
These results indicate that calibration using measured point clouds can reduce the depth discrepancy between the reconstructed geometry and the real environment, thereby providing a more accurate basis for recovering the metric scale of the waypoints.

\begin{table}[t]
\centering
\caption{Quantitative comparison across seven scenarios. Each result is
reported as absolute depth error (m) / relative depth error (\%).
Lower values are better.}
\label{tab:scale_calibration}

\resizebox{\columnwidth}{!}{%
\begin{tabular}{lccc}
\toprule
\textbf{Scenario}
& \textbf{NavDreamer}
& \textbf{DA3}
& \textbf{Ours} \\
\midrule

\textit{Slope}
& 9.521 / 673.01
& 5.170 / 310.34
& \textbf{0.803 / 34.55} \\

\textit{Grassland}
& 3.961 / 345.41
& 4.260 / 331.40
& \textbf{1.211 / 158.41} \\

\textit{KFC}
& 3.353 / 145.57
& 1.760 / 79.68
& \textbf{0.563 / 23.92} \\

\textit{Notice}
& 0.388 / 18.24
& 0.264 / 15.55
& \textbf{0.148 / 14.07} \\

\textit{Red Box 1}
& 0.652 / 32.76
& 1.036 / 40.63
& \textbf{0.170 / 8.21} \\

\textit{Red Box 2}
& 0.407 / 22.82
& 0.583 / 25.56
& \textbf{0.294 / 16.88} \\

\textit{Square}
& 9.662 / 818.66
& 4.048 / 349.51
& \textbf{0.241 / 26.34} \\

\midrule
\textbf{Mean}
& 3.992 / 293.78
& 2.446 / 164.67
& \textbf{0.490 / 40.34} \\

\midrule
\multicolumn{4}{l}{\textit{Mean error reduction achieved by Ours}} \\

vs.\ NavDreamer
& \multicolumn{3}{c}{
\textbf{87.73\% (MADE.) / 86.27\% (MARDE.)}
} \\

vs.\ DA3
& \multicolumn{3}{c}{
\textbf{79.97\% (MADE.) / 75.50\% (MARDE.)}
} \\

\bottomrule
\end{tabular}%
}
\end{table}

\subsection{Qualitative Comparison with a Geometric Planner}
\begin{figure}[h]
    \centering
    \vspace*{2mm}
    \includegraphics[width=0.99\columnwidth]{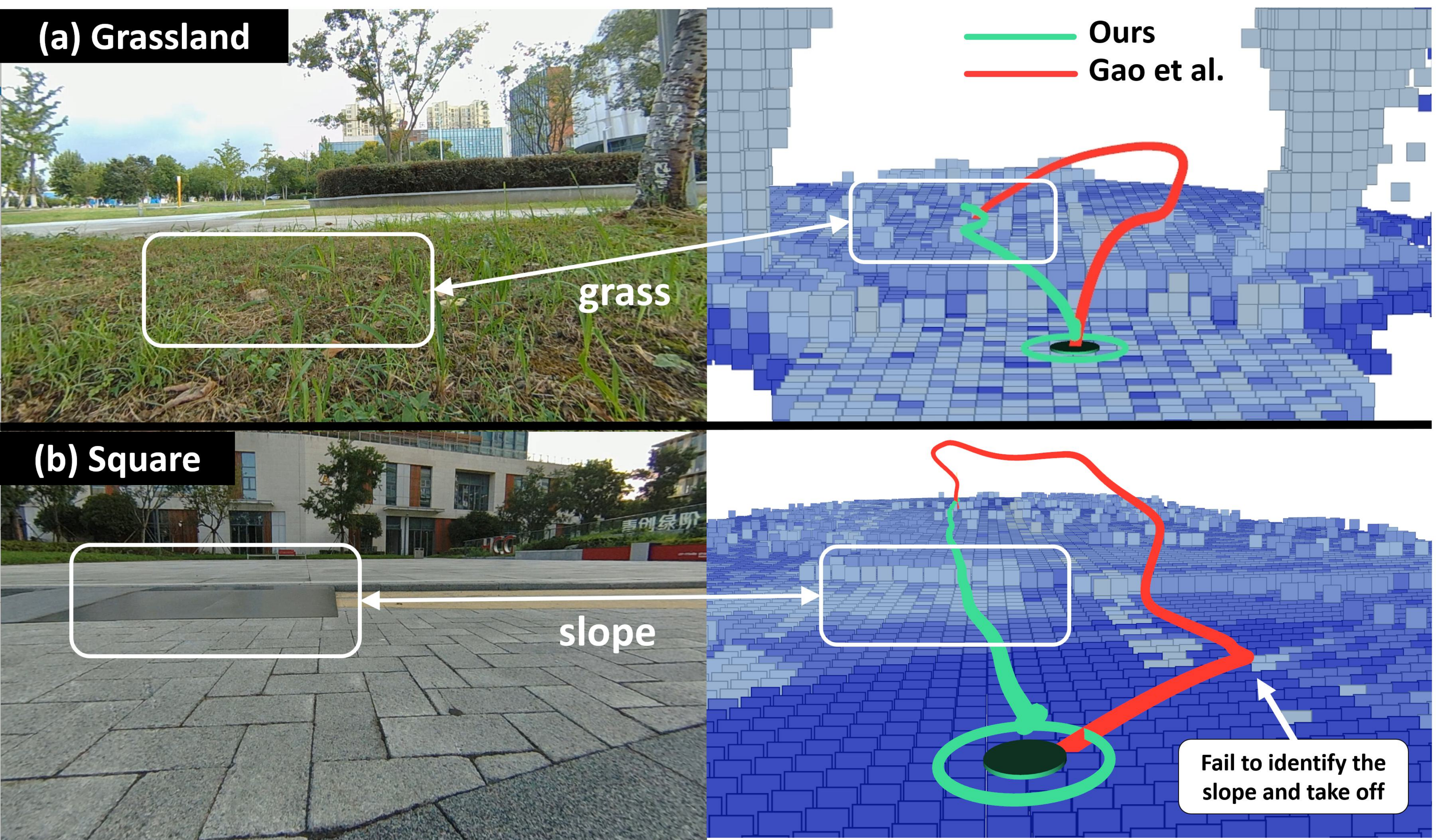}
    \caption{Planned paths in Grassland and Square. Green and red denote TADreamer and a geometric baseline. Ground-extraction and slope-recognition failures trigger baseline flights, while TADreamer selects ground routes using terrain semantics. Insets highlight the relevant terrain.}
    \label{benchmark}
\end{figure}

Fig.~\ref{benchmark} also demonstrates the terrestrial planning capability of our method compared with the planners proposed by Gao et al.~\cite{gao2025autonomous}.
Previous works rely on dedicated ground identification and extraction for terrestrial planning, which fails on uneven terrains.
In the \textit{Grassland} scene, the uneven, grass-covered terrain causes ground extraction to fail, so the planner treats the grass as obstacles and flies along the entire path.
In the \textit{Square} scene, due to the low mapping resolution, the slope cannot be identified, so the planner takes off in front of the step.
In contrast, our planner performs semantic reasoning about ground traversability and therefore selects the ground mode to save energy.

%%%%%%%%%%%%%%%%%%%%%%%%%%%%%%%%%%%%%%%%%%%%%%%%%%%%%%%%%%%%%%%%%%%%%%%%%%%%%%%%
\section{Conclusion}
\label{sec:conclusion}
We presented TADreamer, a language-guided navigation framework that combines video imagination, measured geometric calibration, and mode-aware trajectory planning for terrestrial-aerial bimodal robots. In seven real-world scenarios, the framework produced usable videos within two generation rounds, and all sampled mode annotations in the selected videos agreed with human assessments. Calibration reduced mean absolute depth error by 80.0\% relative to DA3 and 87.7\% relative to NavDreamer. Future work will address high-level closed-loop replanning and navigation in dynamic environments.

%%%%%%%%%%%%%%%%%%%%%%%%%%%%%%%%%%%%%%%%%%%%%%%%%%%%%%%%%%%%%%%%%%%%%%%%%%%%%%%%
% Uncomment this section only when an appendix is needed and permitted.
% \section*{Appendix}

%%%%%%%%%%%%%%%%%%%%%%%%%%%%%%%%%%%%%%%%%%%%%%%%%%%%%%%%%%%%%%%%%%%%%%%%%%%%%%%%
% Uncomment this section for the camera-ready version when appropriate.
% \section*{Acknowledgment}
% Acknowledge assistance and funding here.

% If necessary, use \addtolength{\textheight}{-Xcm} on the page before the last
% page to balance the two columns. Do not use it until the content is final.

\bibliographystyle{IEEEtran}
\bibliography{root}

@inproceedings{hu2025see,
  title={See, point, fly: A learning-free vlm framework for universal unmanned aerial navigation},
  author={Hu, Chih Yao and Lin, Yang-Sen and Lee, Yuna and Su, Chih-Hai and Lee, Jie-Ying and Tsai, Shr-Ruei and Lin, Chin-Yang and Chen, Kuan-Wen and Ke, Tsung-Wei and Liu, Yu-Lun},
  booktitle={Conference on Robot Learning},
  pages={4697--4708},
  year={2025},
  organization={PMLR}
}

@inproceedings{cai2025flightgpt,
  title={Flightgpt: Towards generalizable and interpretable uav vision-and-language navigation with vision-language models},
  author={Cai, Hengxing and Dong, Jinhan and Tan, Jingjun and Deng, Jingcheng and Li, Sihang and Gao, Zhifeng and Wang, Haidong and Su, Zicheng and Sumalee, Agachai and Zhong, Renxin},
  booktitle={Proceedings of the 2025 Conference on Empirical Methods in Natural Language Processing},
  pages={6670--6687},
  year={2025}
}

@article{li2026ds,
  title={DS-LABRNav: Land-Air Bimodal Robot Navigation With Traversable Obstacles Base on Vision-Language Model},
  author={Li, Yongjie and Yu, Wenshuai and Duan, Molong and Zhang, Bo and Liu, Zhou and Li, Qingquan},
  journal={IEEE Robotics and Automation Letters},
  year={2026},
  publisher={IEEE}
}

@article{huang2026navdreamer,
  title={Navdreamer: Video models as zero-shot 3d navigators},
  author={Huang, Xijie and Gai, Weiqi and Wu, Tianyue and Wang, Congyu and Zheng, Qiaoyu and Liu, Zhiyang and Zhou, Xin and Wu, Yuze and Gao, Fei},
  journal={IEEE Robotics and Automation Letters},
  year={2026},
  publisher={IEEE}
}

@inproceedings{fan2019autonomous,
  title={Autonomous hybrid ground/aerial mobility in unknown environments},
  author={Fan, David D and Thakker, Rohan and Bartlett, Tara and Miled, Meriem Ben and Kim, Leon and Theodorou, Evangelos and Agha-mohammadi, Ali-akbar},
  booktitle={2019 IEEE/RSJ International Conference on Intelligent Robots and Systems (IROS)},
  pages={3070--3077},
  year={2019},
  organization={IEEE}
}

@article{zhang2022autonomous,
  title={Autonomous and adaptive navigation for terrestrial-aerial bimodal vehicles},
  author={Zhang, Ruibin and Wu, Yuze and Zhang, Lixian and Xu, Chao and Gao, Fei},
  journal={IEEE Robotics and Automation Letters},
  volume={7},
  number={2},
  pages={3008--3015},
  year={2022},
  publisher={IEEE}
}

@inproceedings{zhang2023model,
  title={Model-based planning and control for terrestrial-aerial bimodal vehicles with passive wheels},
  author={Zhang, Ruibin and Lin, Junxiao and Wu, Yuze and Gao, Yuman and Wang, Chi and Xu, Chao and Cao, Yanjun and Gao, Fei},
  booktitle={2023 IEEE/RSJ International Conference on Intelligent Robots and Systems (IROS)},
  pages={1070--1077},
  year={2023},
  organization={IEEE}
}

@inproceedings{li2025two,
  title={A Two-Stage Lightweight Framework for Efficient Land-Air Bimodal Robot Autonomous Navigation},
  author={Li, Yongjie and Liu, Zhou and Yu, Wenshuai and Lu, Zhangji and Wang, Chenyang and Yu, Fei and Li, Qingquan},
  booktitle={2025 IEEE/RSJ International Conference on Intelligent Robots and Systems (IROS)},
  pages={11165--11171},
  year={2025},
  organization={IEEE}
}

@article{gao2025autonomous,
  title={Autonomous exploration with terrestrial-aerial bimodal vehicles},
  author={Gao, Yuman and Zhang, Ruibin and Lai, Tiancheng and Cao, Yanjun and Xu, Chao and Gao, Fei},
  journal={IEEE Robotics and Automation Letters},
  year={2025},
  publisher={IEEE}
}

@inproceedings{shah2023lm,
  title={Lm-nav: Robotic navigation with large pre-trained models of language, vision, and action},
  author={Shah, Dhruv and Osi{\'n}ski, B{\l}a{\.z}ej and Levine, Sergey and others},
  booktitle={Conference on robot learning},
  pages={492--504},
  year={2023},
  organization={pmlr}
}

@inproceedings{huang2023visual,
  title={Visual language maps for robot navigation},
  author={Huang, Chenguang and Mees, Oier and Zeng, Andy and Burgard, Wolfram},
  booktitle={2023 IEEE International Conference on Robotics and Automation (ICRA)},
  pages={10608--10615},
  year={2023},
  organization={IEEE}
}

@article{du2023learning,
  title={Learning universal policies via text-guided video generation},
  author={Du, Yilun and Yang, Sherry and Dai, Bo and Dai, Hanjun and Nachum, Ofir and Tenenbaum, Josh and Schuurmans, Dale and Abbeel, Pieter},
  journal={Advances in neural information processing systems},
  volume={36},
  pages={9156--9172},
  year={2023}
}

@inproceedings{du2024video,
  title={Video language planning},
  author={Du, Yilun and Yang, Sherry and Florence, Pete and Xia, Fei and Wahid, Ayzaan and Sermanet, Pierre and Yu, Tianhe and Abbeel, Pieter and Tenenbaum, Joshua B and Kaelbling, Leslie and others},
  booktitle={International Conference on Learning Representations},
  volume={2024},
  pages={31138--31155},
  year={2024}
}

@article{chen2026imaginav,
  title={Imaginav: Scalable embodied navigation via generative visual prediction and inverse dynamics},
  author={Chen, Jie and Cai, Yuxin and Wang, Yizhuo and Bai, Ruofei and Cao, Yuhong and Li, Jun and Yun, Yau Wei and Sartoretti, Guillaume},
  journal={arXiv preprint arXiv:2603.13833},
  year={2026}
}

@article{sam2026action,
  title={Action Agent: Agentic Video Generation Meets Flow-Constrained Diffusion},
  author={Sam, Jeffrin and Khang, Nguyen and Mahmoud, Yara and Cabrera, Miguel Altamirano and Tsetserukou, Dzmitry},
  journal={arXiv preprint arXiv:2605.01477},
  year={2026}
}

@article{lin2025depth,
  title={Depth anything 3: Recovering the visual space from any views},
  author={Lin, Haotong and Chen, Sili and Liew, Junhao and Chen, Donny Y and Li, Zhenyu and Shi, Guang and Feng, Jiashi and Kang, Bingyi},
  journal={arXiv preprint arXiv:2511.10647},
  year={2025}
}

@article{li2026triphibot,
  title={TriphiBot: a triphibious robot combining FOC-based propulsion with eccentric design},
  author={Li, Xiangyu and Lai, Tiancheng and Lai, Mingwei and Lin, Junxiao and Zhang, Mengke and Zhi, Junping and Xu, Chao and Gao, Fei and Cao, Yanjun},
  journal={arXiv preprint arXiv:2602.01385},
  year={2026}
}

@inproceedings{besl1992method,
  title={Method for registration of 3-D shapes},
  author={Besl, Paul J and McKay, Neil D},
  booktitle={Sensor fusion IV: control paradigms and data structures},
  volume={1611},
  pages={586--606},
  year={1992},
  organization={Spie}
}

@inproceedings{lai2025trofybot,
  title={TrofyBot: A Transformable Rolling and Flying Robot with High Energy Efficiency},
  author={Lai, Mingwei and Ye, Yuqian and Wu, Hanyu and Xuan, Chice and Zhang, Ruibin and Ren, Qiuyu and Xu, Chao and Gao, Fei and Cao, Yanjun},
  booktitle={2025 IEEE International Conference on Robotics and Automation (ICRA)},
  pages={4989--4995},
  year={2025},
  organization={IEEE}
}

@article{lin2024skater,
  title={Skater: A novel bi-modal bi-copter robot for adaptive locomotion in air and diverse terrain},
  author={Lin, Junxiao and Zhang, Ruibin and Pan, Neng and Xu, Chao and Gao, Fei},
  journal={IEEE Robotics and Automation Letters},
  volume={9},
  number={7},
  pages={6392--6399},
  year={2024},
  publisher={IEEE}
}

@article{serpiva2026dreamtonav,
  title={DreamToNav: Generalizable Navigation for Robots via Generative Video Planning},
  author={Serpiva, Valerii and Sam, Jeffrin and Simon, Chidera and Amjad, Hajira and Zhura, Iana and Lykov, Artem and Tsetserukou, Dzmitry},
  journal={arXiv preprint arXiv:2603.06190},
  year={2026}
}

@article{du2010scaling,
  title={Scaling iterative closest point algorithm for registration of m--D point sets},
  author={Du, Shaoyi and Zheng, Nanning and Xiong, Lei and Ying, Shihui and Xue, Jianru},
  journal={Journal of Visual Communication and Image Representation},
  volume={21},
  number={5-6},
  pages={442--452},
  year={2010},
  publisher={Elsevier}
}

@inproceedings{yokoyama2024vlfm,
  title={Vlfm: Vision-language frontier maps for zero-shot semantic navigation},
  author={Yokoyama, Naoki and Ha, Sehoon and Batra, Dhruv and Wang, Jiuguang and Bucher, Bernadette},
  booktitle={2024 IEEE International Conference on Robotics and Automation (ICRA)},
  pages={42--48},
  year={2024},
  organization={IEEE}
}

@article{hu2024metric3d,
  title={Metric3d v2: A versatile monocular geometric foundation model for zero-shot metric depth and surface normal estimation},
  author={Hu, Mu and Yin, Wei and Zhang, Chi and Cai, Zhipeng and Long, Xiaoxiao and Chen, Hao and Wang, Kaixuan and Yu, Gang and Shen, Chunhua and Shen, Shaojie},
  journal={IEEE Transactions on Pattern Analysis and Machine Intelligence},
  volume={46},
  number={12},
  pages={10579--10596},
  year={2024},
  publisher={IEEE}
}

@inproceedings{wang2026pi,
  title={{$\pi^3$}: Permutation-Equivariant Visual Geometry Learning},
  author={Wang, Yifan and Zhou, Jianjun and Zhu, Haoyi and Chang, Wenzheng and Zhou, Yang and Li, Zizun and Chen, Junyi and Pang, Jiangmiao and Shen, Chunhua and He, Tong},
  booktitle={International Conference on Learning Representations},
  volume={2026},
  pages={10481--10497},
  year={2026}
}

@article{wang2026moge,
  title={Moge-2: Accurate monocular geometry with metric scale and sharp details},
  author={Wang, Ruicheng and Xu, Sicheng and Dong, Yue and Deng, Yu and Xiang, Jianfeng and Lv, Zelong and Sun, Guangzhong and Tong, Xin and Yang, Jiaolong},
  journal={Advances in Neural Information Processing Systems},
  volume={38},
  pages={35928--35959},
  year={2026}
}

@article{lai2026explore,
  title={Explore From Sketch: Accelerating UAV Exploration in Large-scale Environments with Prior Maps},
  author={Lai, Tiancheng and Gao, Yuman and Li, Xiangyu and Pang, Ruitian and Wang, Xingpeng and Shen, Siqi and Zhang, Mengke and He, Yin and Gao, Fei and Xu, Chao and others},
  journal={arXiv preprint arXiv:2606.11708},
  year={2026}
}

\end{document}